\documentclass[]{rsos}

\usepackage{times}
\usepackage{epsfig}
\usepackage{graphicx}
\usepackage{amsmath}
\usepackage{amssymb}

\usepackage{multirow}
\usepackage{url}
\usepackage{threeparttable}

\begin{document}

\title{Objective Classes for Micro-Facial Expression Recognition}

\author{
Adrian K. Davison$^{1}$, Walied Merghani$^{2}$ and Moi Hoon Yap$^{3}$}

\address{$^{1}$Centre for Imaging Sciences, University of Manchester, Manchester, United Kingdom\\
$^{2}$Sudan University of Science and Technology, Khartoum, Sudan\\
$^{3}$School of Computing, Mathematics and Digital Technology, Manchester Metropolitan University, Manchester, United Kingdom\\}

\subject{computer vision, pattern recognition, feature descriptor}


\corres{Moi Hoon Yap\\
\email{M.Yap@mmu.ac.uk}}

\begin{abstract}
Micro-expressions are brief spontaneous facial expressions that appear on a face when a person conceals an emotion, making them different to normal facial expressions in subtlety and duration. Currently, emotion classes within the CASME II dataset are based on Action Units and self-reports, creating conflicts during machine learning training. We will show that classifying expressions using Action Units, instead of predicted emotion, removes the potential bias of human reporting. The proposed classes are tested using LBP-TOP, HOOF and HOG 3D feature descriptors. The experiments are evaluated on two benchmark FACS coded datasets: CASME II and SAMM. The best result achieves 86.35\% accuracy when classifying the proposed 5 classes on CASME II using HOG 3D, outperforming the result of the state-of-the-art 5-class emotional-based classification in CASME II. Results indicate that classification based on Action Units provides an objective method to improve micro-expression recognition.
\end{abstract}


\begin{fmtext}
\section{Introduction}

A micro-facial expression is revealed when someone attempts to conceal their true emotion~\cite{Ek04,Ek09}. When they consciously realise that a facial expression is occurring, the person may try to suppress the facial expression because showing the emotion may not be appropriate~\cite{Ma08b}. Once the suppression has occurred, the person may mask over the original facial expression and cause a micro-facial expression. In a high-stakes environment, these expressions tend to become more likely as there is more risk to showing the emotion.

\end{fmtext}

\maketitle

The duration of a micro-expression is very short and is considered the main feature that distinguishes them from a facial expression~\cite{Sh12}, with the general standard being a duration of no more than 500 ms~\cite{Ya13a}. Other definitions of speed that have been studied show micro-expressions to last less than 200 ms, this defined by Ekman and Friesen~\cite{Ek69} as first to describe a micro-expression, 250 ms~\cite{Ek01}, less than 330 ms~\cite{Ek05} and less than half a second~\cite{Fr09b}. 

Micro-facial expression analysis is less established and harder to implement due to being less distinct than normal facial expressions. Feature representations, such as Local Binary Patterns (LBP)~\cite{Oj96,Oj02,Zh07a}, Histogram of Oriented Gradients (HOG)~\cite{Da05} and Histograms of Oriented Optical Flow (HOOF)~\cite{Ch09}, are commonly used to describe micro-expressions. Although micro-facial expression analysis is very difficult, the popularity in recent years has grown due to the potential applications in security and interrogations~\cite{Os09,Fr09b,Fr09a,yap2013database}, healthcare~\cite{Ho92,Co09} and automatic detection in real-world applications where the detection accuracy of humans peaks around 40\%~\cite{Fr09b}.

Generally, the process of recognising normal facial expressions involves preprocessing, feature extraction and classification. Micro-expression recognition is not an exception, but the features extracted should be more descriptive due the small movement in micro-expressions compared with normal expressions. One of the biggest problems faced by research in this area is the lack of publicly available datasets, which the success in facial expression recognition~\cite{Ya14d} research largely relies on. Gradually, datasets of spontaneously induced micro-expression have been developed~\cite{Li13a,Ya13b,Ya14a,Da16a}, but earlier research was centred around posed datasets~\cite{Po09,Sh11}.

Eliciting spontaneous micro-expression is a real challenge because it can be very difficult to induce the emotions in participants and also get them to conceal them effectively in a lab-controlled environment. Micro-expression datasets need decent ground truth labelling with Action Units (AUs) using the Facial Action Coding System (FACS)~\cite{Ek78a}. FACS objectively assigns AUs to the muscle movements of the face. If any classification of movements take place for micro-facial expressions, it should be done with AUs and not only emotions. Emotion classification requires the context of the situation for an interpreter to make a meaningful interpretation. Most spontaneous micro-expression datasets have FACS ground truth labels and estimated or predicted emotion. These have been annotated by an expert and self-reports written by participants. 

We contend that using AUs to classify micro-expressions gives more accurate results than using predicted emotion categories. By organising the AUs of the two most recent FACS coded state-of-the-art datasets, CASME II~\cite{Ya14a} and SAMM~\cite{Da16a}, into objective classes, we ensure that the learning methods train on specific muscle movement patterns and therefore increase accuracy. Yan et al.~\cite{Ya14c} also state that it's inappropriate to categorise micro-expressions into emotion categories, and that using FACS AU research to inform the eventual emotional classification.

To date, experiments on micro-expression recognition using categories based purely on AU movements, has not been completed. Additionally, the SAMM dataset was designed for micro-movement analysis rather than recognition. We contribute by completing recognition experiments on the SAMM dataset for the first time with three features previously used for micro-expression analysis: LBP-TOP~\cite{Zh07a}, HOOF~\cite{Ch09} and HOG 3D~\cite{Da05,Po09}. Further, the proposed objective classes could inform future research on the importance of objectifying movements of the face.

The remainder of this paper is divided into the following sections; Section 2 discusses the background of two of the FACS coded state-of-the-art datasets developed for micro-expression analysis and the related work in micro-expression recognition; Section 3 describes the methodology; Section 4 presents the results and discusses the effects of applying objective classification to a micro-expression recognition task; Section 6 concludes this paper and discusses future work.

\section{Background}
This section will describe two datasets which are used in the experiments for this paper. A comparative summary of the datasets can be seen in Table~\ref{tab:datasetSum}. Previously developed micro-expression recognition systems are also discussed using established features to represent each micro-expression.
\begin{table}[h]
	\centering
	\begin{threeparttable}
		
		\caption{A summary of the different features of the CASME II and SAMM datasets.}
		\label{tab:datasetSum}
		\renewcommand{\arraystretch}{1.3}
		\tabcolsep=0.05cm
		\begin{tabular}{|l|l|l|}
			\hline
			Feature & CASME II~\cite{Ya14a} & SAMM~\cite{Da16a} \\ \hline
			Micro-Movements & 247\tnote{*} & 159 \\ \hline
			Participants & 35 & 32 \\ \hline
			Resolution & 640$\times$480 & 2040$\times$1088 \\ \hline
			Facial Resolution & 280$\times$340 & 400$\times$400 \\ \hline
			FPS & 200 & 200 \\ \hline
			Spontaneous/Posed & Spontaneous & Spontaneous \\ \hline
			FACS Coded & Yes & Yes \\ \hline
			No. Coders & 2 & 3 \\ \hline
			Emotion Classes & 5 & 7 \\ \hline
			Mean Age (SD) & 22.03 (SD = 1.60) & 33.24 (SD = 11.32) \\ \hline
			Ethnicities & 1 & 13 \\ \hline
		\end{tabular}
		\begin{tablenotes}\footnotesize
			\item[*] This is the original amount of movements used in \cite{Ya14a}, however we use a larger set of 255 provided by the dataset.
		\end{tablenotes}
	\end{threeparttable}
\end{table}
\subsection{CASME II}
CASME II was developed by Yan et al.~\cite{Ya14a} and refers to Chinese Academy of Sciences Micro-expression Database II, which was preceded by CASME~\cite{Ya13b} with major improvements. All samples in CASME II are spontaneous and dynamic micro-expressions with a high frame rate (200 fps). There are a few frames kept before and after each micro-expression to make it suitable for detection experiments. The resolution of samples is 640$\times$480 pixels for recording which saved as MJPEG and resolution about 280$\times$340 pixels for cropped facial area. The participants' facial expressions were elicited in a well-controlled laboratory environment.
%
%
The dataset contains 255 micro-expressions (gathered from 35 participants) and were selected from nearly 3000 facial movements and have been labelled with AUs based on FACS. Only 247 movements were used in the original experiments on CASME II~\cite{Ya14a}. The inter-coder reliability of the FACS codes within the dataset is 0.846. Flickering light was avoided in the recordings and highlights to regions of the face was reduced. However, there were some limitations: firstly, the materials used for eliciting micro-expression are video episodes which can have different meanings to different people, for example eating worms may not always disgust someone. Secondly, micro-expressions are elicited under one specific lab situation. There was some difficulty in eliciting some types of facial expressions in laboratory situations, such as sadness.

\begin{figure*}
	\centering
	\includegraphics[scale=0.43]{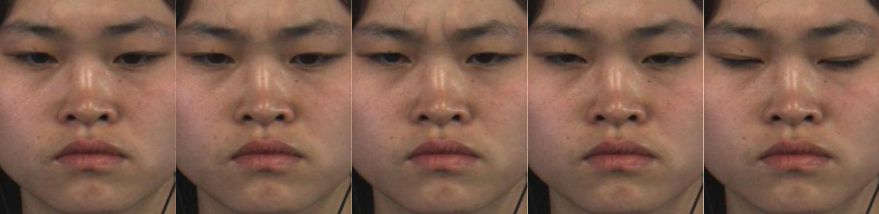}
	\caption{Sample frames showing Subject 11's micro-expression clip \textquoteleft EP19\_03f\textquoteright \ that was coded as an AU4 in the \textquoteleft others\textquoteright \ category.}
	\label{fig:sub11}
\end{figure*}
\begin{figure*}
	\centering
	\includegraphics[scale=0.4]{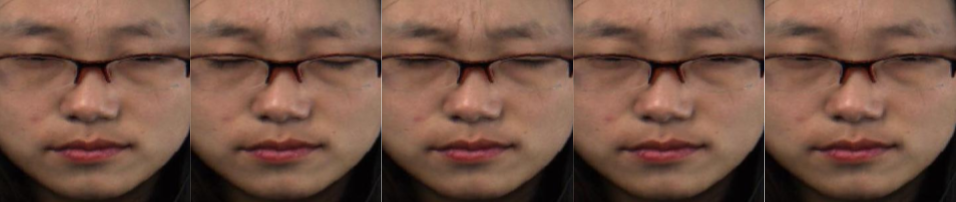}
	\caption{Sample frames showing Subject 26's micro-expression clip \textquoteleft EP18\_50\textquoteright \ that was coded as an AU4 in the \textquoteleft disgust\textquoteright \ category.}
	\label{fig:sub26}
\end{figure*}
When analysing the FACS codes of the CASME II dataset, it was found that there are many conflicts to the coded AUs and the estimated emotions. These inconsistencies do not help when attempting to train distinct machine learning classes, and adds further justification for the proposed introduction of new classes based on AUs only.

For example, Subject 11 with the micro-expression clip filename of \textquoteleft EP19\_03f\textquoteright, was coded as an AU4 in the \textquoteleft others\textquoteright \, estimated emotion category (shown in Fig.~\ref{fig:sub11}). However, Subject 26 with the micro-expression clip filename of \textquoteleft EP18\_50\textquoteright, was also coded with AU4 but in the \textquoteleft disgust\textquoteright \, estimated emotion category (shown in Fig.~\ref{fig:sub26}). As can be seen in the apex frame (centre image) of both Fig.~\ref{fig:sub11} and ~\ref{fig:sub26}, AU4, the lowering of the brow, is present. Having the same movement in different categories is likely to have an effect on any training stage of machine learning.

\subsection{SAMM}
The Spontaneous Actions and Micro-Movements (SAMM)~\cite{Da16a} dataset is the first high-resolution dataset of 159 micro-movements induced spontaneously with the largest variability in demographics. To obtain a wide variety of emotional responses, the dataset was created to be as diverse as possible. A total of 32 participants were recruited for the experiment with a mean age of 33.24 years (SD: 11.32, ages between 19 and 57), and an even gender split of 16 male and female participants. The inter-coder reliability of the FACS codes within the dataset is 0.82, and was calculated by using a slightly modified version of the inter-reliability formula, found in the FACS Investigator's Guide~\cite{Ek78b}, to account for three coders rather than two.

The inducement procedure was based on the 7 basic emotions~\cite{Ek04} and recorded at 200 fps. As part of the experimental design, each video stimuli was tailored to each participant, rather than obtaining self reports during or after the experiment. This allowed for particular videos to be chosen and shown to participants for optimal inducement potential. The experiment comprised of 7 stimuli used to induce emotion in the participants who were told to suppress their emotions so that micro-facial movements might occur. To increase the chance of this happening, a prize of \pounds50 was offered to the participant that could hide their emotion the best, therefore introducing a high-stakes situation~\cite{Ek04,Ek09}. Each participant completed a questionnaire prior to the experiment so that the stimuli could be tailored to each individual to increase the chances of emotional arousal.

The SAMM dataset was originally designed to investigate micro-facial movements by analysing muscle movements of the face rather than recognising distinct classes~\cite{Da15}. We are the first to categorise SAMM based on the FACS AUs and then use these categories for micro-facial expression recognition.
\subsection{Related Work}
Currently, there are three features which many micro-expression recognition approaches rely on: Local Binary Patterns (LBP), Histogram of Oriented Gradients (HOG) and Histogram of Oriented Optical Flow (HOOF) based. We will discuss different methods that use these features in recent work on micro-expression recognition. Further, specific important micro-expression research is discussed.

As an extension to the original Local Binary Pattern (LBP)~\cite{Oj02} operator, Local Binary Patterns on Three Orthogonal Planes (LBP-TOP) was proposed by Zhao et al.~\cite{Zh07a} demonstrated to be effective for dynamic texture and facial expression analysis in the spatial-temporal domain. Since video sequence of time length T, usually it can be thought as a stack of XY planes along the time axis T, but also it can be thought as three planes XY, XT and YT. These provide information about space and time transition. The basic idea of LBP-TOP is similar to LBP, the difference being that LBP-TOP extracts features from all three planes which will be combined in into a single feature vector.

Yan et al.~\cite{Ya14a} carried out the first micro-expression recognition experiment on the CASME II dataset. LBP-TOP~\cite{Zh07a} was used to extract the features and Support Vector Machine (SVM)~\cite{Co95b} was employed as the classifier. The radii varied from 1 to 4 for X and Y, and from 2 to 4 for T (T=1 was not considered due to little change between two neighbouring frames at 200 fps), with classification occurring between five main categories of emotions provided in this experiment (happiness, disgust, surprise, repression and others).

Davison et al.~\cite{Da14} used the LBP-TOP feature to differentiate between movements and neutral sequences, attempting to avoid bias when classifying with an SVM.

The performance of~\cite{Ya14a} on recognising micro-expressions in 5-classes with LBP-TOP features extraction, achieved a best result of 63.41\% accuracy, using leave-one-out cross-validation. This result is an average for recent micro-expression recognition research, and is likely due to the way micro-expressions are categorised. Of the 5-class in the CASME II dataset, 102 were classed as \textquoteleft others\textquoteright, which denoted movements not suited for the other categories but related to emotion. The next highest category was \textquoteleft disgust\textquoteright \, with 60 movements, showing that the \textquoteleft others\textquoteright \, class made the categorisation imbalanced. Further, the categorisation was not based solely on AUs due to micro-expressions being short in duration and low in intensity, but also based on the participant's self-reporting. By classifying micro-expressions in this way, features are unlikely to exhibit a pattern, and therefore perform poorly during the recognition stage as can be seen in the other performance results. For example, in ~\cite{Ya14a}, the highest results is 63.41\%, which is still relatively low.

More recently, LBP-TOP was used as a base feature for micro-expression recognition with integral projection~\cite{Hu15a,Hu16b}. These representations attempt to improve discrimination between micro-expression classes and therefore improve recognition rates.
Polikovsky et al.~\cite{Po09} used a 3D gradient histogram descriptor (HOG 3D) to recognise posed micro-facial expressions from high-speed videos. The paper used manually marked up areas that are relevant to FACS-based movement so that unnecessary parts of the face are left out. This does means that the method of classifying movement in these subjectively selected areas is time-consuming and would not suit a real-time application like interrogation. The spatio-temporal domain is explored highlighting the importance of the temporal plane in micro-expressions, however the bin selection for the XY plane is 8 and the XT, YT planes have been set to 12. The number of bins selected represents the different directions of movement in each plane.

For HOOF-based methods, a Main Direction Mean Optical Flow (MDMO) feature was proposed by Liu et al.~\cite{Li15a} for micro-facial expression recognition using SVM as a classifier. The method of detection also uses 36 regions, partitioned using 66 facial points on the face, to isolate local areas for analysis, but keeping the feature vector small for computational efficiency. The best result on the CASME II dataset was 67.37\% using leave-one-subject-out cross validation.

The basic HOOF descriptor was also used by Li et al.~\cite{Li15b} as a comparative feature when spotting micro-expressions and then performing recognition. This is the first automatic micro-expression system which can spot and recognise micro-expressions from spontaneous video data, and be comparable to human performance.

Using Robust Principal Component Analysis (RPCA)~\cite{Wr09}, Wang et al.~\cite{Wa14d} extract the sparse information from micro-expression data, and then use Local Spatiotemporal Directional Features, based on LBP-TOPs dynamic features, to extract the subtle motion on the face from 16 ROIs of local importance to facial expression motion.

A novel colour space model was created named Tensor Independent Color Space (TICS), that helps recognise micro-expressions~\cite{Wa15b}. By extracting the LBP-TOP features of independent colour components, micro-expression clips can be better recognised than the RGB space.

Huang et al.~\cite{Hu16a} proposed Spatio-Temporal Completed Local Quantization Patterns (STCLQP), which extracts the sign, magnitude and orientation of the micro-expression data, then an efficient vector quantization and codebook selection are developed in both the appearance
and temporal domains for generalising classical pattern types. Finally, using the developed codebooks, spatio-temporal features of sign, magnitude and orientation components are extracted and fused, with experiments being run on SMIC, CASME and CASME II.

By exploiting the sparsity in the spatial and temporal domains of micro-expressions, a Sparse Tensor Canonical Correlation Analysis was proposed for micro-expression characteristics~\cite{Wa16}. This method reduces the dimensionality of micro-expression data and enhances LBP coding to find a subspace to maximise the correlation between micro-expression data and their corresponding LBP code.

Liong et al.~\cite{Li16} investigate the use of only two frames from a micro-expression clip: the onset and the apex frame. By only using a couple of frames, a good accuracy is achieved when using the proposed Bi-Weighted Oriented Optical Flow feature to encode the expressiveness of the apex frame.

As micro-movements on the face are heavily affected by the global movements of a person's head, Xu et al.~\cite{Xu17} propose a Facial Dynamics Map to distinguish between what is a micro-expression and what would be classed as a non-micro-expression. The facial surface movement between adjacent frames is predicted using optical flow. The movements are then extracted in a coarse-to-fine manner, indicating different levels of facial dynamics. This step is used to differentiate micro-movements from anything else. Finally, an SVM is used for both identification and categorisation.

Wang et al.~\cite{Wa17} recently proposed a Main Directional Maximal Difference (MDMD) method that uses the magnitude maximal difference in the main direction of optical flow features to find when facial movements occur. These movements can be used for both micro-expressions and macro-expressions to find the onset, apex and offset of a movement within the context of each examined clip.
\section{Methodology}

To overcome the conflicting classes in CASME II, we restructure the classes around the AUs that have been FACS coded. Using EMFACS~\cite{Ek78b}, a list of AUs and combinations are proposed for a fair categorisation of the SAMM~\cite{Da16a} and CASME II~\cite{Ya14a} datasets. Categorising in this way removes the bias of human reporting and relies on the ground truth movement data, feature representation and recognition technique for each micro-expression clip. Table~\ref{tab:AUCat} shows 7 classes and the corresponding AUs that have been assigned to that class. Classes I-VI are linked with happiness, surprise, anger, disgust, sadness and fear. Class VII relates to contempt and other AUs that have no emotional link in EMFACS~\cite{Ek78b}. It should be noted that the classes do not directly correlate to being these emotions, however the links used are informed from previous research~\cite{Ek76,Ek78a,Ek78b}.
\begin{table}
	\centering
	\caption{Each class represents AUs that can be linked to emotion.}
	\label{tab:AUCat}
	\renewcommand{\arraystretch}{1.2}
		\begin{tabular}{|c|l|}
			\hline
			\multicolumn{1}{|l|}{Class} & Action Units \\ \hline
			I & AU6, AU12, AU6+AU12, AU6+AU7+AU12, AU7+AU12 \\ \hline
			II & \begin{tabular}[c]{@{}l@{}}AU1+AU2, AU5, AU25, AU1+AU2+AU25, AU25+AU26,\\ AU5+AU24\end{tabular} \\ \hline
			III & \begin{tabular}[c]{@{}l@{}}A23, AU4, AU4+AU7, AU4+AU5, AU4+AU5+AU7,\\ AU17+AU24, AU4+AU6+AU7, AU4+AU38\end{tabular} \\ \hline
			IV & \begin{tabular}[c]{@{}l@{}}AU10, AU9, AU4+AU9, AU4+AU40, AU4+AU5+AU40,\\ AU4+AU7+AU9, AU4 +AU9+AU17, AU4+AU7+AU10,\\ AU4+AU5+AU7+AU9, AU7+AU10\end{tabular} \\ \hline
			V & AU1, AU15, AU1+AU4, AU6+AU15, AU15+AU17 \\ \hline
			VI & AU1+AU2+AU4, AU20 \\ \hline
			VII & Others \\ \hline
	\end{tabular}
\end{table}
Each movement in both datasets were classified based on the AU categories of Table~\ref{tab:AUCat}, with the resulting frequency of movements being shown in Table~\ref{tab:compClassDatasets}.
\begin{table}[h]
	\centering
	\caption{The total number of movements assigned to the new classes for both SAMM and CASME II.}
	\label{tab:compClassDatasets}
	\renewcommand{\arraystretch}{1.2}
	\begin{tabular}{ | c | c | c | c | }
		\hline
		Class & CASME II & SAMM & Total \\ \hline
		I & 25 & 24 & 49 \\ \hline
		II & 15 & 13 & 28 \\ \hline
		III & 99 & 20 & 119 \\ \hline
		IV & 26 & 8 & 34 \\ \hline
		V & 20 & 3 & 23 \\ \hline
		VI & 1 & 7 & 8 \\ \hline
		VII & 69 & 84 & 153 \\ \hline
		Total & 255 & 159 & 415 \\ \hline
	\end{tabular}
\end{table}

Micro-expression recognition experiments are run on two datasets: CASME II and SAMM. For this experiment, three types of feature representations are extracted from a sequence of grey images which represent the micro-expressions. These image sequences are divided into 5$\times$5 blocks that are non-overlapping. The LBP-TOP features~\cite{Zh07a} radii parameters for X, Y and T are set to 1, 1 and 4 respectively and all neighbours in three planes set to 4. The HOG 3D~\cite{Po09} and HOOF~\cite{Ch09} features are set to the parameters described in the original implementations.

Sequential Minimal Optimization (SMO)~\cite{Pl99} is used in the classification phase with 10-fold cross validation and leave-one-subject-out (LOSO) to classify between I-V, I-VI and I-VII classes. SMO is a fast algorithm for training SVMs, and provide a solution to solving very large quadratic programming (QP) problems, which are required to train SVMs. SMO avoid time-consuming QP calculations by breaking them down into smaller pieces. Doing this allows for the classification task to be completed much faster than using traditional SVMs~\cite{Pl99}.
\section{Results}
\begin{figure}
	\centering
	\includegraphics[scale=0.92]{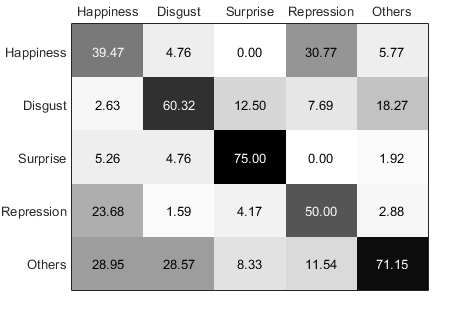}
	\caption{Confusion matrix of the original CASME II classes using the LBP-TOP feature, using SMO as a classifier.}
	\label{fig:ConfMatOrigCASMEII}
\end{figure}
\begin{figure}
	\centering
	\includegraphics[scale=0.92]{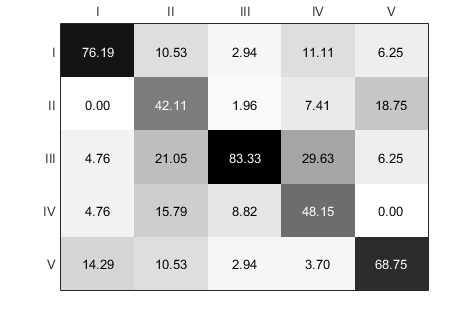}
	\caption{Confusion matrix of the proposed classes I-V on the CASME II dataset using the LBP-TOP feature and SMO as a classifier.}
	\label{fig:confMat_newClass}
\end{figure}
Evidence to support the proposed AU-based categories can be seen in the confusion matrix in Fig.~\ref{fig:ConfMatOrigCASMEII}. A high proportion of micro-expressions have been classified as \textquoteleft others\textquoteright, for example 28.95\% of the \textquoteleft happiness\textquoteright \,and 28.57\% of the \textquoteleft disgust\textquoteright \,categories are classified as \textquoteleft others\textquoteright \,respectively. The original chosen emotions, including many placed in the \textquoteleft others\textquoteright \, category, leads to a lot of conflict at the recognition stage. It should be noted that the CASME II dataset~\cite{Ya14a} included self-reporting, which adds another layer of complexity during classification.

The proposed classes I-V classification results using LBP-TOP can be seen in the confusion matrix in Fig.~\ref{fig:confMat_newClass}. In contrast, the classification rates are more stable and outperforming the original classes overall. The results are by no means perfect, however it shows that the most logical direction is to use objective classes based on AUs rather than estimated emotion categories. Further investigation using an objective selection of FACS-based regions~\cite{Da16b} supports this with AUC results for detecting relevant movements to be 0.7512 and 0.7261 on SAMM and CASME II, respectively.

\begin{table*}
	\centering
	\caption{Results on the CASME II dataset showing each feature, proposed classes, and the original classes defined in~\cite{Ya14a} for comparison.}
	\label{tab:CASMEII_Res}
	\renewcommand{\arraystretch}{1.2}
	\resizebox{1\textwidth}{!}{%
		\begin{tabular}{|c|c|c|c|c|c|c||c|c|c|c|c|}
			\hline
			&  & \multicolumn{5}{c||}{10-Fold Cross-Validation} & \multicolumn{5}{c|}{Leave-One-Subject-Out (LOSO)} \\ \hline
			Feature & Class & Accuracy (\%) & TPR & FPR & F-Measure & AUC & Accuracy (\%) & TPR & FPR & F-Measure & AUC \\ \hline
			\multirow{4}{*}{LBP-TOP} & Original & 77.17 & 0.56 & 0.22 & 0.53 & 0.74 & 68.24 & 0.49 & 0.17 & 0.48 & 0.63 \\ \cline{2-12} 
			& I-V & 77.94 & 0.63 & 0.33 & 0.58 & 0.70 & 67.80 & 0.54 & 0.14 & 0.51 & 0.44 \\ \cline{2-12} 
			& I-VI & 76.84 & 0.59 & 0.32 & 0.55 & 0.69 & 67.94 & 0.53 & 0.14 & 0.51 & 0.44 \\ \cline{2-12} 
			& I-VII & 76.13 & 0.50 & 0.23 & 0.45 & 0.70 & 61.92 & 0.39 & 0.17 & 0.35 & 0.63 \\ \hline
			\multirow{4}{*}{HOOF} & Original & 78.83 & 0.61 & 0.19 & 0.60 & 0.78 & 68.36 & 0.51 & 0.24 & 0.49 & 0.61 \\ \cline{2-12} 
			& I-V & 82.70 & 0.69 & 0.22 & 0.67 & 0.80 & 69.64 & 0.59 & 0.18 & 0.56 & 0.47 \\ \cline{2-12} 
			& I-VI & 82.41 & 0.68 & 0.23 & 0.66 & 0.79 & 73.52 & \textbf{0.62} & 0.18 & \textbf{0.60} & 0.47 \\ \cline{2-12} 
			& I-VII & 83.94 & 0.64 & 0.14 & 0.63 & 0.79 & \textbf{76.60} & 0.57 & \textbf{0.14} & 0.55 & \textbf{0.72} \\ \hline
			\multirow{4}{*}{HOG3D} & Original & 80.93 & 0.62 & 0.14 & 0.62 & 0.79 & 59.59 & 0.38 & 0.24 & 0.35 & 0.50 \\ \cline{2-12} 
			& I-V & \textbf{86.35} & \textbf{0.72} & 0.13 & \textbf{0.72} & \textbf{0.84} & 69.53 & 0.56 & 0.18 & 0.51 & 0.40 \\ \cline{2-12} 
			& I-VI & 83.49 & 0.68 & 0.16 & 0.67 & 0.80 & 69.87 & 0.56 & 0.18 & 0.51 & 0.40 \\ \cline{2-12} 
			& I-VII & 82.59 & 0.58 & \textbf{0.12} & 0.58 & 0.79 & 61.33 & 0.39 & 0.30 & 0.31 & 0.51 \\ \hline
	\end{tabular}}
\end{table*}

\begin{table*}
	\centering
	\caption{Results on the SAMM dataset showing each feature and proposed classes.}
	\label{tab:SAMM_Res}
	\renewcommand{\arraystretch}{1.2}
	\resizebox{1\textwidth}{!}{%
		\begin{tabular}{|c|c|c|c|c|c|c||c|c|c|c|c|}
			\hline
			&  & \multicolumn{5}{c||}{10-Fold Cross-Validation} & \multicolumn{5}{c|}{Leave-One-Subject-Out (LOSO)} \\ \hline
			Feature & Class & Accuracy (\%) & TPR & FPR & F-Measure & AUC & Accuracy (\%) & TPR & FPR & F-Measure & AUC \\ \hline
			\multirow{3}{*}{LBP-TOP} & I-V & 79.21 & 0.54 & 0.16 & 0.51 & 0.74 & 44.70 & 0.38 & 0.19 & 0.35 & 0.31 \\ \cline{2-12} 
			& I-VI & \textbf{81.93} & 0.55 & \textbf{0.13} & 0.52 & \textbf{0.74} & 45.89 & 0.34 & 0.17 & 0.31 & 0.36 \\ \cline{2-12} 
			& I-VII & 79.52 & 0.57 & 0.18 & \textbf{0.56} & 0.74 & 54.93 & 0.42 & 0.22 & 0.39 & \textbf{0.40} \\ \hline
			\multirow{3}{*}{HOOF} & I-V & 78.95 & 0.56 & 0.16 & 0.55 & 0.74 & 42.17 & 0.32 & \textbf{0.06} & 0.33 & 0.32 \\ \cline{2-12} 
			& I-VI & 79.53 & 0.52 & 0.15 & 0.51 & 0.73 & 40.89 & 0.28 & 0.07 & 0.27 & 0.35 \\ \cline{2-12} 
			& I-VII & 72.80 & 0.52 & 0.32 & 0.50 & 0.65 & 60.06 & 0.49 & 0.25 & \textbf{0.48} & 0.30 \\ \hline
			\multirow{3}{*}{HOG3D} & I-V & 77.18 & 0.51 & 0.17 & 0.49 & 0.74 & 34.16 & 0.22 & 0.15 & 0.22 & 0.24 \\ \cline{2-12} 
			& I-VI & 79.41 & 0.48 & 0.15 & 0.45 & 0.71 & 36.39 & 0.19 & 0.14 & 0.19 & 0.26 \\ \cline{2-12} 
			& I-VII & 79.09 & \textbf{0.59} & 0.25 & 0.55 & 0.71 & \textbf{63.93} & \textbf{0.50} & 0.22 & 0.44 & 0.30 \\ \hline
	\end{tabular}}
\end{table*}

Table~\ref{tab:CASMEII_Res} shows the experimental results on CASME II with each result metric being a weighted average calculation to account for imbalanced numbers within classes. Each experiment was completed for each feature and within the original classes defined in~\cite{Ya14a} and the proposed classes. Both the 10-fold cross-validation results and leave-one-subject-out (LOSO) are shown.

The top performing feature achieves a weighted accuracy score of 86.35\% for the HOG 3D feature in the proposed class I-V. This shows a large improvement over the original classes which achieved 80.93\% for the same feature. Using LOSO, the results were comparable with the original classes. The highest accuracy was 76.60\% from the HOOF feature, in the proposed I-VII classes. For the CASME II dataset results, using LBP-TOP and 10-fold cross-validation, the original method outperformed the classes I-VI and I-VII. In addition, for HOG3D LOSO, the original method outperforms in class I-VII when using F-measure as a measurement.

The experiment based on the same conditions were then repeated for SAMM and can be seen in Table~\ref{tab:SAMM_Res}. Overall the recognition rates were good for SAMM, with the best result achieving an accuracy of 81.93\% using LBP-TOP in I-VI classes for 10-fold cross validation. The best result using LOSO was from the HOG 3D feature, in the proposed I-VII classes and achieved 63.93\%, however due to the lower amount of micro-expressions within the SAMM dataset compared with CASME II, the LOSO results were lower.

Some results show that using LOSO, HOOF outperforms in CASME II while HOG3D outperforms in SAMM and in CASME II using LOSO, the HOOF feature achieves a higher accuracy for classes I-VII over I-VI, but not for the F-measure metric. Explanations of this comes down to the data, and how large some variations of the settings, such as resolution and capture methods, are set. The imbalance of data, specifically the low amounts of micro-expression data, can skew LOSO results with low amounts of testing and training. This shows how using LOSO for micro-expression recognition is difficult to quantify with a fair amount of significance. Further data collection of spontaneous micro-expressions is required to rectify this.
%
\section{Conclusion}
We show that restructuring micro-expression classes objectively around the AUs, recognition results outperform the state-of-the-art, emotion-based classification approaches. As micro-expressions are so subtle, the best way to categorise is objectively as possible, so using AU codes is the most logical. Categorising using a combination of AUs and self-reports~\cite{Ya14a} can cause many conflicts when training a machine learning method. Further, dataset imbalances can be very detrimental to machine learning algorithms, and this is further emphasised with the relatively low amount of movements in both datasets. Future work will look into the effect of using more modern features, with AUs classification to improve on the recognition accuracy. This could include the MDMO feature~\cite{Li15a}, local wrinkle feature \cite{ng2015wrinkle} and the feature extraction methods described by Wang et al.~\cite{Wa15a}.

Further work can be done to improve micro-facial expression datasets. Firstly, more datasets or expanding previous sets would be a simple improvement that can help move the research forward faster. Secondly, a standard procedure on how to maximise the amount of micro-movements induced spontaneously in laboratory controlled experiments would be beneficial. If collaboration between established datasets and researchers from psychology occurred, dataset creation would be more consistent.

Deep learning has emerged as a new area of machine learning research~\cite{Be09,De14,Al17}, and micro-expression analysis has yet to exploit this trend. Unfortunately, the amount of high-quality spontaneous micro-expression data is low and deep learning requires a large amount of data to work well~\cite{De14}. Many video-based datasets previously used have over 10,000 video samples~\cite{So12} and even over 1 million actions extracted from YouTube videos~\cite{Ka14}. A real effort to gather spontaneous micro-expression data is required for deep learning approaches to be effective in the future.

\vskip1pc


\dataccess{Data available from: http://www2.docm.mmu.ac.uk/STAFF/m.yap/dataset.php}

\aucontribute{A.K. Davison carried out the design of the study, the re-classification of the Action Units grouping and drafted the manuscript (The tasks was completed when A.K. Davison was in Manchester Metropolitan University). W. Merghani conducted the experiments, analysed the data and drafted the manuscript. M.H.Yap designed the study, developed the theory, assisted development and testing and edited the manuscript. All the authors have read and approved this version of the manuscript.}

\competing{We declare we have no competing interests.}

\funding{This work was completed in Manchester Metropolitan University on a "Future Research Leaders Programme" awarded to M.H. Yap. M.H.Yap is a Royal Society Industry Fellow. }












\bibliographystyle{RS} 
\bibliography{completeBibliography} 

\end{document}